\pdfoutput=1

\documentclass[11pt]{article}

\usepackage[final]{acl}
\usepackage{ellipsis}
\usepackage{amsthm}
\usepackage{amsmath}
\usepackage{algorithmic}
\usepackage{svg}
\usepackage{graphicx}
\usepackage{amssymb}
\usepackage{pifont}

\usepackage{times}
\usepackage{booktabs}
\usepackage{tabularx}

\usepackage{blindtext}

\usepackage{latexsym}
\usepackage[T1]{fontenc}
\usepackage{tabularray}
\usepackage[utf8]{inputenc}
\usepackage{microtype}
\usepackage{multirow}
\usepackage{multicol}
\usepackage{mathrsfs}
\usepackage{array}
\usepackage{booktabs}

\usepackage{todonotes}

\usepackage{algorithm}
\usepackage{algorithmic}
\usepackage{soul}

\usepackage[final]{pdfpages}

\title{Agreement Tracking for Multi-Issue Negotiation Dialogues}

\author{Amogh Mannekote \\
  University of Florida \\
  \texttt{amogh.mannekote@ufl.edu} \\\And
  Bonnie J. Dorr \\
  University of Florida \\
  \texttt{bonniejdorr@ufl.edu} \\\And 
  Kristy Elizabeth Boyer \\
  University of Florida \\
  \texttt{keboyer@ufl.edu}
  }

\author{Amogh Mannekote,$^1$ 
        Bonnie J. Dorr,$^1$
        Kristy Elizabeth Boyer$^1$ \AND
\textnormal{$^1$University of Florida} \AND
Correspondence: {\tt \{\href{mailto:amogh.mannekote@nyu.edu}{amogh.mannekote},
\href{mailto:bonniejdorr@ufl.edu}{bonniejdorr},
\href{mailto:keboyer@ufl.edu}{keboyer}\}@ufl.edu}}

\begin{document}
\maketitle
\begin{abstract}
Automated negotiation support systems aim to help human negotiators reach more favorable outcomes in multi-issue negotiations (e.g., an employer and a candidate negotiating over issues such as salary, hours, and promotions before a job offer). To be successful, these systems must accurately track agreements reached by participants in real-time.
Existing approaches either focus on task-oriented dialogues or produce unstructured outputs, rendering them unsuitable for this objective. Our work introduces the novel task of agreement tracking for two-party multi-issue negotiations, which requires continuous monitoring of agreements within a structured state space.
To address the scarcity of annotated corpora with realistic multi-issue negotiation dialogues, we use GPT-3 to build \textsc{GPT-Negochat}, a synthesized dataset that we make publicly available. We present a strong initial baseline for our task by transfer-learning a T5 model trained on the MultiWOZ 2.4 corpus. Pre-training T5-small and T5-base on MultiWOZ 2.4's DST task enhances results by 21\% and 9\% respectively over training solely on \textsc{GPT-Negochat}. We validate our method's sample-efficiency via smaller training subset experiments. By releasing \textsc{GPT-Negochat} and our baseline models, we aim to encourage further research in multi-issue negotiation dialogue agreement tracking.
\end{abstract}

\section{Introduction}
Negotiation dialogues are common in both adversarial and collaborative contexts. However, a long line of foundational research in psychology and business has established that in general, humans tend to be poor negotiators, often failing to maximize favorable outcomes. Consequently, developing capabilities to build automated systems that can support human negotiators has been an active area of research \cite{prakken2006formal, wang2019persuasionforgood}.

We focus specifically on multi-issue negotiation dialogues,\footnote{Also called ``integrative'' negotiation \cite{zhan2022letsnegotiate}.} where participants negotiate over more than one issue, e.g., salary and hours for a job, or location and time for a meeting. These type of dialogues arise in a wide range of real-life situations that span the adversarial-cooperative spectrum. For example, most job negotiations fall on the adversarial end of the spectrum, since the two parties often have opposing objectives. On the other hand, planning-based meetings are characterized as more cooperative in that participants collaborate on a shared plan to achieve a common goal. Regardless of the specific real-world context for the multi-issue negotiation, the ability to track agreements in real-time is a mission-critical capability for any system aiming to effectively support the participants.

Previous work has examined related tasks such as dialogue summarization and action-item generation at the end of a dialogue. Our work is the first to investigate the task of agreement tracking for multi-issue negotiation dialogues over a structured state space (i.e. an ``ontology''). Agreement tracking 
requires reasoning over multiple turns in a dialogue, in contrast to single utterance-level natural language understanding (NLU).

Agreement tracking also differs from dialogue state tracking (DST)  in that it necessitates explicit agreement from both interlocutors in contrast to DST’s estimation of the goals of a single interlocutor. However, it is not at all obvious how one leverages existing task formulations (including DST), model designs, and training objectives for agreement tracking without manually collecting and labelling a substantial amount of in-domain data. We demonstrate a deficiency in existing methods for three closely-related tasks: building negotiation dialogue agents, summarization of meetings, and dialogue state tracking for task-oriented dialogues.

\begin{enumerate}
    \item \textbf{Negotiation Dialogue Agents:} Existing research on automated negotiation systems focuses on strategic aspects, e.g., undervaluing or appealing to the opposing party \cite{zhou2019adynamicstrategy, he2018decouplingstrategy, keizer2017evaluatingpersuasion}. This is orthogonal to the problem of tracking agreements. End-to-end negotiators, which directly generate responses \cite{he2017learningsymmetric}, do not generate intermediate structured representations of the dialogue state. A limited number of studies have investigated language understanding in this context, but have done so only at the level of single utterances \cite{chawla2021casinoacorpus, yamaguchi2021dialogueactbased, frampton2009realtimedecision}. Agreement tracking, in contrast, necessitates reasoning across multiple turns since acceptances and rejections can only be understood in reference to offers made in previous turns.
    
    \item \textbf{Meeting Summarization:} The release of the AMI Meeting Corpus \cite{kraaij2005ami} has stimulated significant research on meeting and dialogue summarization \cite{gliwa2019samsumcorpusa, wang2013domainindependent, rennard2022abstractivemeeting, kryscinski2020evaluatingthefactual, liu2019automaticdialogue}. However, meeting summarization differs from agreement tracking in two crucial ways. First, summarization is applied at the end of a dialogue, whereas tracking takes place continuously. Second, summarization focuses on generating a human-readable output of the key decisions made during the meetings, whereas the tracking task generates a structured representation of the agreements based on a fixed ontology, and this representation serves as input to downstream modules.
    
    \item \textbf{Dialogue State Tracking (DST):} The goal of DST in task-oriented dialogue is to extract users' goals by inferring the values for a predefined set of keys (commonly referred to as ``slots''). As we shall see later, we re-purpose this notion of a slot to track agreement over a single issue in our multi-issue negotiation dialogue setup. Dialogue state tracking has been a long-standing task in task-oriented dialogue systems literature \cite{Williams2016TheDS, jacqmin2022doyoufollow, zhao2021effectivesequence, Rastogi2019TowardsSM}. Although there has been extensive work in recent years to improve the ability of task-oriented DST models to generalize to unseen domains with zero-shot \cite{lin2021leveragingslotdescriptions, campagna2020zeroshottransfer} and few-shot \cite{Wu2019TransferableMS} models, these are still limited to form-filling dialogues (e.g., restaurant reservation and hotel booking). Moreover, these models are challenged by agreement tracking, as it requires explicit agreement from both interlocutors, unlike DST that estimate the goals of a single interlocutor. Hence, the question of how effectively state-of-the-art DST techniques can be used for state tracking in vastly different dialogue paradigms (such as negotiation) remains open-ended.
\end{enumerate}

To establish a strong foundation for future modeling efforts in agreement tracking, we introduce a transfer-learning approach that involves pre-training a T5 model \cite{raffel2020exploringthelimits} using the MultiWOZ 2.4 corpus. Subsequently, we fine-tune the model on the \textsc{GPT-Negochat} corpus, which we introduce in Section \ref{sec:dataset}, for the agreement tracking task. Our approach outperforms a T5 model that is fine-tuned only on \textsc{GPT-Negochat} (without the MultiWOZ corpus). Additionally, we investigate the sample-efficiency of our model in low-resource settings by experimenting with various fractional splits of our training data. Sample efficiency (ability 
to learn from a small number of demonstrations) is a key concern for our task since it is expensive to collect a lot of annotated dialogue samples for a new domain. We assess slot prediction accuracy of all models using Joint Slot Accuracy and Joint F1 Score. Our findings illustrate that incorporating a pre-training phase, using T5-small and T5-base on MultiWOZ 2.4's DST task, improves performance by 21\% and 9\% respectively (in absolute percentage points), compared to exclusively training on \textsc{GPT-Negochat}.

The rest of the paper is organized as follows.
Section \ref{sec:dataset}
describes the creation and characteristics of \textsc{GPT-Negochat}, the dataset used for our experiments. 
Section \ref{sec:preliminaries} 
formally defines
the task of agreement tracking,
the tokenization scheme, the training objective, and the evaluation metrics used in this study. Sections \ref{sec:experiments} and \ref{sec:results} describe our experimental procedure and results, respectively. Section \ref{sec:related} presents the related work. Finally, Section \ref{sec:limitations} talks about the limitations of our work. 

\section{GPT-Negochat: A Multi-Issue Negotiation Dialogue Corpus}
\label{sec:dataset}

The corpus selected for our experiments must satisfy two criteria: inclusion of multi-issue negotiation dialogues and availability of ground-truth annotations for agreements at each dialogue turn. Two publicly-available corpora meet these requirements: the \textsc{Negochat} corpus \cite{konovalov_negochat_2016} and the \textsc{Metalogue} corpus for multi-issue bargaining dialogues \cite{Petukhova2016ModellingMB}. However, we ruled out \textsc{Metalogue} (due to paywall restrictions) and conduct experiments solely on the \textsc{Negochat} corpus, which is freely accessible and imposes no copyright limitations on usage.

The \textsc{Negochat} corpus \cite{konovalov_negochat_2016} contains 105 crowd-sourced dialogues (1484 utterances) between an Employer and a Candidate who negotiate over issues surrounding a job offer such as salary, role, and working hours (the complete ontology of the \textsc{Negochat} corpus is outlined in Table \ref{tab:negochat-ontology}). While the Employer-side of the conversations is supplied by human participants on Amazon M-Turk,\footnote{https://www.mturk.com/} an automated agent plays the role of the Candidate. A second human ``wizard'' is responsible for acting as a live Natural Language Understanding (NLU) module that parses the Employers' utterances into a structured semantic representation. This parsed input is then processed by an automated dialogue manager, which generates a response using a template-based NLG module.

\begin{table}[ht]
\resizebox{\columnwidth}{!}{
\begin{tabular}{@{}p{2.5cm}p{5.8cm}@{}}
\toprule
\textbf{Slot Type} & \textbf{Possible Values} \\ \midrule
Working Hours & 8 hours, 9 hours, 10 hours \\ \hline
Pension Fund & 10\%, 20\% \\ \hline
Job Description & Programmer, Team Manager, Project Manager \\ \hline
Promotion Possibilities & Slow promotion track, Fast promotion track \\ \hline
Salary & 90k USD, 60k USD, 120k USD \\ \hline
Leased Car & With leased car, Without leased car, No agreement \\ 
\bottomrule
\end{tabular}
}
\caption{Ontology of the GPT-Negochat corpus.\label{tab:negochat-ontology}}
\end{table}

Due to limited lexical diversity in the Candidate-side utterances of the \textsc{Negochat} corpus (see Table \ref{tab:nego-vs-gpt}), attributed to its basic template-based NLG module, we employed GPT-3 to enrich and naturalize the corpus \cite{brown2020languagemodelsare}. We used prompt templates (Appendix \ref{sec:gpt-negochat-prompt}) to rephrase the utterances, with the revised version's efficacy evaluated by two sets of human annotations on a random 10\% dataset slice, adjudicating any discrepancies.

Out of the utterance pairs presented to the annotators, the rephrased utterances of \textsc{GPT-Negochat} were reported as being more realistic in 78.4\% of the cases, and in 7\% of the cases, both the original and rephrased utterances were rated as equally realistic. The utterance pairs were deemed to carry the same meaning in the context of the previous turn in 91.4\% of the annotated instances, attesting to the effectiveness of the updated \textsc{GPT-Negochat} corpus. The full text of the instructions presented to the annotators are shown in Appendix \ref{sec:human-annotation-instructions}.

Despite being smaller than some multi-domain DST datasets, the resulting dataset provides a valuable benchmark for agreement tracking models in negotiations, especially taking into consideration that constructing a high-quality, labeled dataset is timeconsuming. Dialogue comparisons between the original \textsc{Negochat} and \textsc{GPT-Negochat} are provided in Table \ref{tab:nego-vs-gpt} and Appendix \ref{sec:gpt-neg-vs-neg-examples}. We name this revised version of the \textsc{Negochat} corpus as 
``\textsc{GPT-Negochat}'' and make it publicly available for use by the broader research community.\footnote{\href{Hidden for review}{Hidden for review}}

\begin{table*}[ht]
\centering
\resizebox{0.98\textwidth}{!}{
\small
\begin{tabular}{@{}p{2cm}p{6cm}p{6cm}@{}}
\toprule
\textbf{Speaker} & \textbf{Original Utterance} & \textbf{Rephrased Utterance} \\ \midrule
Candidate & I would like a position of project manager & I'm interested in a position as a project manager. \\ \hline
Employer & are you sure you wouldnt rather be a programmer? & Are you sure that's the job you're looking for? Wouldn't you prefer to be a programmer? \\ \hline
Candidate & I refuse programmer position. I am expecting a position of project manager & I'm sorry, but I'm not interested in the programmer position. I'm looking for a project manager role instead. \\ \hline
Employer & what about Quality assurance? & What about a Quality Assurance role? \\  \hline
Candidate & I reject qa position. I would like a position of project manager & No, thank you. I'm only interested in a project manager position. \\
\bottomrule
\end{tabular}
}
\caption{An excerpt from a dialogue from the \textsc{Negochat} corpus alongside its rephrased counterpart from \textsc{GPT-Negochat}.\label{tab:nego-vs-gpt}}
\end{table*}

\section{Preliminaries} \label{sec:preliminaries}
This section provides formal definitions for notations and describes the input and output representations used in our model.

\subsection{Notations}
\label{sec:notations}

We consider a negotiation dialogue between two participants taking alternating turns with utterances $\{T_1, \dots, T_N\}$. To maintain a strict alternating turn structure, consecutive utterances made by the same participant are merged into a single utterance; thus, for each $T_i$ the odd $i$'s represent one speaker, and the even $i$'s represent the other. The negotiation focuses on a predefined set of issues, i.e., the domain's ``ontology,'' in accordance with the DST framework. Each issue is associated with one of $M$ slots
$\{s_1, \dots, s_M\}$ from our ontology, for example slot $s_1$ could refer to Salary, slot $s_2$ to Working Hours, and so on. 

At each dialogue turn $t$, the agreement state $A_t$ is defined as a slot-value mapping between each issue in the ontology and the corresponding value that both participants have agreed upon for it. We denote this slot-value relationship in the agreement state as $A_t(s_j) = v$ (where $v=\epsilon$ indicates that no agreement has been reached thus far).

Finally, we associate each utterance $T_t$ with a list of (one or more) dialogue acts, $D_t = [d_t^1, \dots, d_t^{|d_t|}]$, where $d_t^i$ is one of $\{\textsc{Offer}, \textsc{Accept}, \textsc{Reject}, \textsc{Other}\}$. These dialogue acts, defined as stated below, are important inputs for our rule-based algorithm defined in Section~\ref{sec:baselines}. 

\begin{enumerate}
    \item \textsc{Offer}: When making an offer, a participant puts forth one or more slot-value pairs for consideration by the other participant. We formally represent an offer as a list of these key-value pairs: $O(\{(s_1, v_1), \dots, (s_{n_i}, v_{n_i}))\}$. We constrain each issue to a single slot value.
    \item \textsc{Accept} and \textsc{Reject}: While accepting or rejecting an offer, a participant can do so either partially or completely. As an example of partial acceptance, if the Candidate demands an 8-hour workday with a pension of 20\%, the Employer might choose to accept the 8-hour workday, but not the 20\% pension. 
    \item \textsc{Other}: Utterances of other types (e.g., greetings) are categorized under the \textsc{Other} label as they have no direct effect on the agreement state. 
\end{enumerate}

\begin{table*}[t]
\centering
\small
\label{tab:sample-dialogue}
\begin{tabular}{l|l} 
\toprule
\multicolumn{1}{c|}{\textbf{Conversation}} & \multicolumn{1}{c}{\textbf{Agreement State}} \\ 
\toprule
\textbf{E}: No company car included? & No agreements \\ 
\hline
\textbf{C}: Right, no car. Let's move on. I was hoping for a pension of 20\%. & \textbf{Company Car}: No \\ 
\hline
\textbf{E}: If you work 10 hours, I can offer you a 20\% pension - what do you think? & \textbf{Company Car}: No \\ 
\hline
\textbf{C}: No thanks. I'm expecting an 8-hour workday and I want a 10\% pension & \textbf{Company Car}: No \\ 
\hline
\textbf{E}: How about a salary of 60K if you agree to the 10\% pension? & \textbf{Company Car}: No \\ 
\hline
\textbf{C}: I'm sorry, but I'm looking for a salary of 120,000 and a pension of 20\%. & \textbf{Company Car}: No \\ 
\hline
\textbf{E}: What about offering you a fast promotion tract with a 90k salary? & \textbf{Company Car}: No \\ 
\hline
\textbf{C}: No, I'm afraid that won't work for me. & \textbf{Company Car}: No \\ 
\hline
\textbf{E}: Would you be comfortable with a salary of 60 or 90k? & \textbf{Company Car}: No \\ 
\hline
\textbf{C}: 90,000 sounds good to me. Is there anything else we need to discuss? & \begin{tabular}[c]{@{}l@{}}\textbf{Company Car}: No \\ \textbf{Salary:} 90,000\end{tabular} \\
\bottomrule
\end{tabular}
\caption{An excerpt of a dialogue from \textsc{GPT-Negochat} with turn-level agreement annotations. \textbf{C} and \textbf{E} stand for Candidate and Employer respectively.}
\vspace{3em}
\end{table*}

\subsection{Representing Dialogue Context}
Transformer-based generative language models, such as those introduced by \citet{Vaswani2017AttentionIA}, have commonly been used for representing dialogue context in dialogue tasks. This typically involves concatenating utterances within a context window. Dialogue state tracking (DST) models have followed this practice.

Full-history based DST models consider the entire dialogue up to a turn as their context \cite{HosseiniAsl2020ASL, Feng2020ASA, Peng2021SoloistBT}. However, as the number of utterances increases beyond a certain point, these models struggle to retain information. Recursive approaches to DST combat this issue by considering a small window of utterances (typically one to four) as their dialogue context \cite{Lin2020MinTLMT, Budzianowski2019HelloIG, Lei2018SequicityST}. Rather than predicting the entire dialogue state from scratch at every turn, these approaches use the previously predicted dialogue state as a starting point. For our experiments, we opt for a recursive-based approach due to the relatively high average turn-length of 34.27 \textsc{GPT-Negochat}, which exceeds that of most task-oriented dialogue corpora.

\subsection{Levenshtein Belief Spans}
In task-oriented dialogue, a \textit{belief} refers to the dialogue system's understanding of the user's goal at a given dialogue turn. This belief is continuously updated based on the user's incoming utterances. A \textit{belief span} is a text-based representation of the system's belief. For example, a belief span such as ``cuisine = Chinese; area = center'' may indicate that a user is seeking a downtown Chinese restaurant. Recent DST approaches adopt conditional text generation models to generate belief spans corresponding to the updated dialogue state. This allows for easier expansion to new slot types and values without complete model retraining.

We employ the Levenshtein Belief Spans (Levs) tokenization method  \citet{Lin2020MinTLMT}. This approach, which we term \textit{A-Lev}, combines a domain prefix (either [gpt-negochat] or [multiwoz]) with a sequence of edit operations.  Edit operations include insertion, deletion, and substitution. Although \textsc{GPT-Negochat} focuses on a single domain (job offer negotiation), we apply a common domain prefix across all samples. A more detailed explanation of the \textit{Lev} construction method can be found in the original paper \citet{Lin2020MinTLMT}. Figure \ref{fig:prompts} shows a tokenized training example used to train our agreement tracking models.

\section{Experiments} \label{sec:experiments}

We present a rule-based reference model that serves as the basis for our main results. We then introduce our experimental setup, including our choice of backbone model, rule-based baseline model, training objectives, and evaluation metrics.

\subsection{Backbone Model (T5)}
We choose T5 \cite{raffel2020exploringthelimits} as our backbone model. T5 is a text-to-text Transformer model consisting of a multi-layer encoder and decoder pre-trained on autoregressive language modeling, text denoising, and deshuffling objectives. Its text-based input-output representation enables broad NLP task applicability without architectural changes. Table \ref{tab:t5-stats} enumerates the parameters, layers, embedding dimension, and attention heads for each T5 variant. However, due to computational constraints, we restrict our experiments to the T5-small and T5-base versions. While we recognize the benefits of exploring other LLMs, our focus presently is to set T5 as a benchmark for this task, intending to evaluate and compare with other models in future research.

\begin{table}[ht]
\small
\makebox[\columnwidth]{
\begin{tabular}{@{}lllll@{}}
\toprule
\textbf{Model} & \textbf{Parameters} & \textbf{\# layers} & \textbf{d\_model} & \textbf{\# heads} \\
\midrule
Small          & 60M                 & 6                  & 512               & 8                 \\
Base           & 220M                & 12                 & 768               & 12                \\
\bottomrule
\end{tabular}
}
\caption{Properties of the Small and Base variants of the T5 model. $d\_model$ and \# heads refer to the dimensionality of the model's hidden state and the number of attention heads respectively.\label{tab:t5-stats}}
\end{table}

\begin{figure*}[th]
    \centering
    \includegraphics[trim=0cm 8cm 1cm 8cm, scale=0.45]{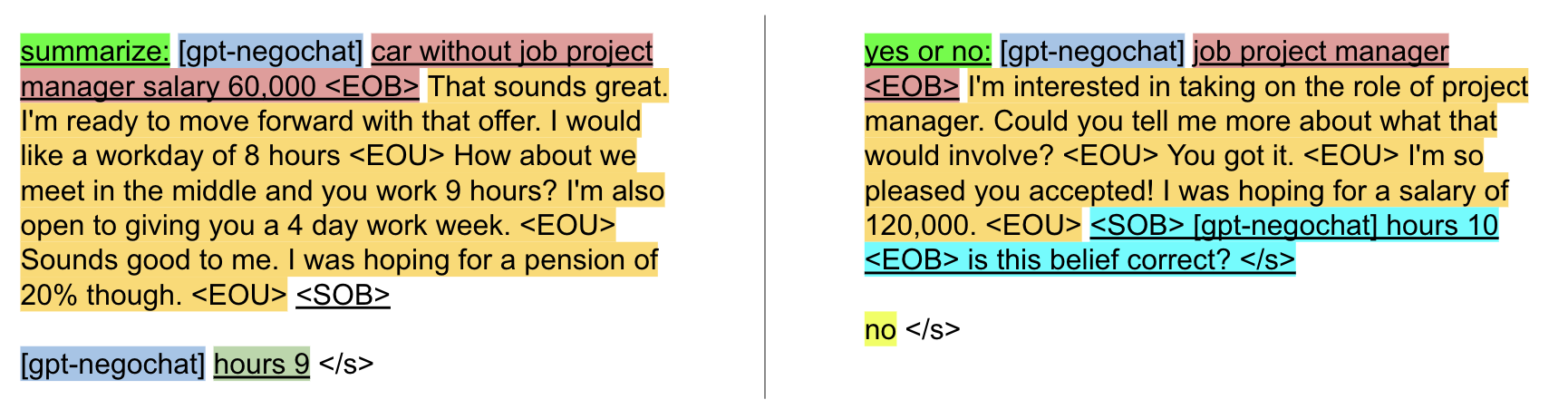}
    \vspace{85pt}
    \caption{This figure shows the tokenization schemes for both \textsc{Gen} (left) and the \textsc{Clf} (right) tasks. The \textsc{Gen} task input includes task prefix (green), dataset prefix (blue), belief span representation (red), and concatenated utterances within a window size (yellow). Its output contains the dataset prefix (blue) and the updated Levenshtein belief span (dark green). The input in the \textsc{Clf} scheme is mostly identical to that of \textsc{Gen}, except for its unique task-prefix and an appended candidate belief span. Its output is simply ``yes'' or a ``no''. Alternating spans are underlined for visual clarity.}
    \label{fig:prompts}
\end{figure*}

\subsection{Rule-Based Agreement Tracker}
\label{sec:baselines}


\begin{figure}[H]
    \centering
    \includegraphics[width=\columnwidth]{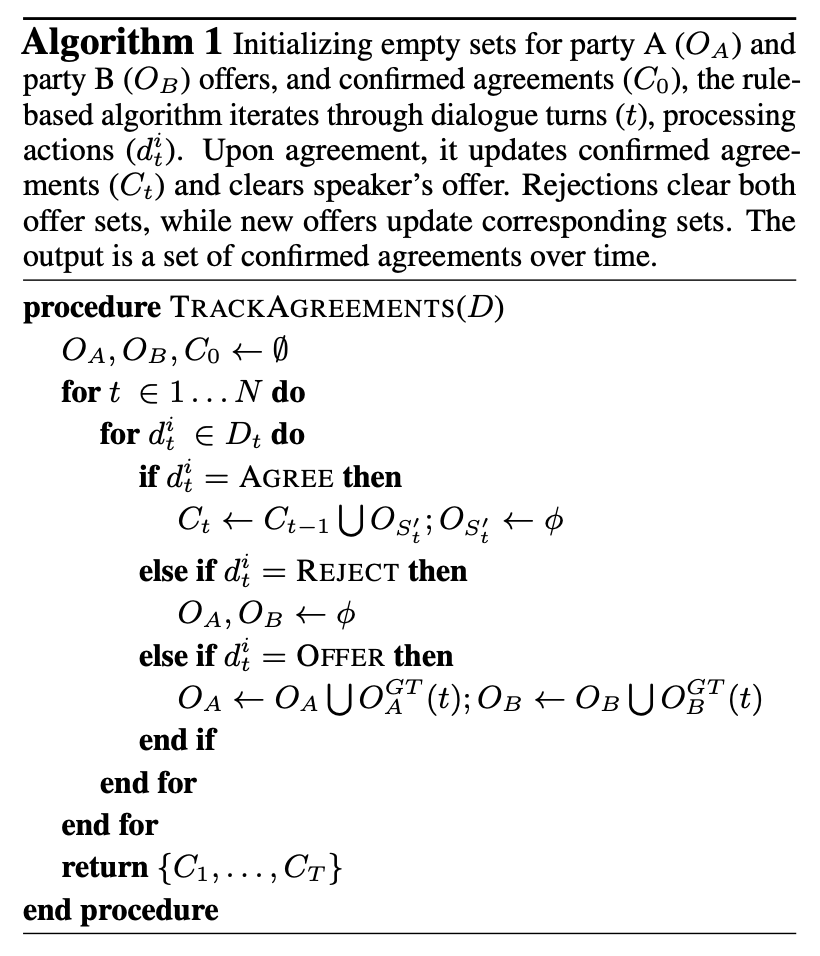}
    \label{alg:baseline}
\end{figure}

We develop a rule-based reference model, \textsc{Oracle-Tracker}, with oracle access to turn-level ground truth annotations to contextualize our transfer-learning model's results and assess the multi-turn reasoning difficulty in agreement tracking. This model focuses on tracking agreements across multiple turns, assuming known dialogue acts (Offer, Accept, Reject, and Other) and entities. 

The model addresses two subtasks: (1) natural language understanding (NLU) at the utterance level, where it classifies dialogue acts and extracts slot-value pairs; and (2) tracking agreements over multiple turns. In designing this rule-based reference model, we focus on the demonstrably more difficult second task that involves multi-turn reasoning as described below.

To quantify the relative simplicity of the single-turn NLU problem, we train a T5-base model on just 10 \textsc{GPT-Negochat} dialogues to classify the dialogue acts and extract corresponding slot-value pairs from each utterance. The model achieves a correct prediction rate of 74.33\% on the test set. In contrast, our top-performing model for agreement tracking, also trained on 10 dialogues, manages only a 20\% joint slot accuracy (refer to Section \ref{sec:results}).

\subsection{Transfer Learning from Dialogue State Tracking}
Tracking information across dialogue turns is a shared goal of agreement tracking and dialogue state tracking, despite differing focus areas (agreements vs user goals). To leverage the abundance of publicly annotated data for dialogue state tracking, we choose to pre-train our model on this task. This pre-training equips the model with transferable representations for this general tracking problem, enabling efficient adaptation to agreement tracking. In our experiments, we compare joint slot accuracy results of two T5 models to assess transfer efficiency: one solely fine-tuned for agreement tracking, the other sequentially fine-tuned for dialogue state tracking and agreement tracking.

    
    
    
    
     
     
    
    
     

\begin{figure*}
    \centering
    \includegraphics[scale=0.41]{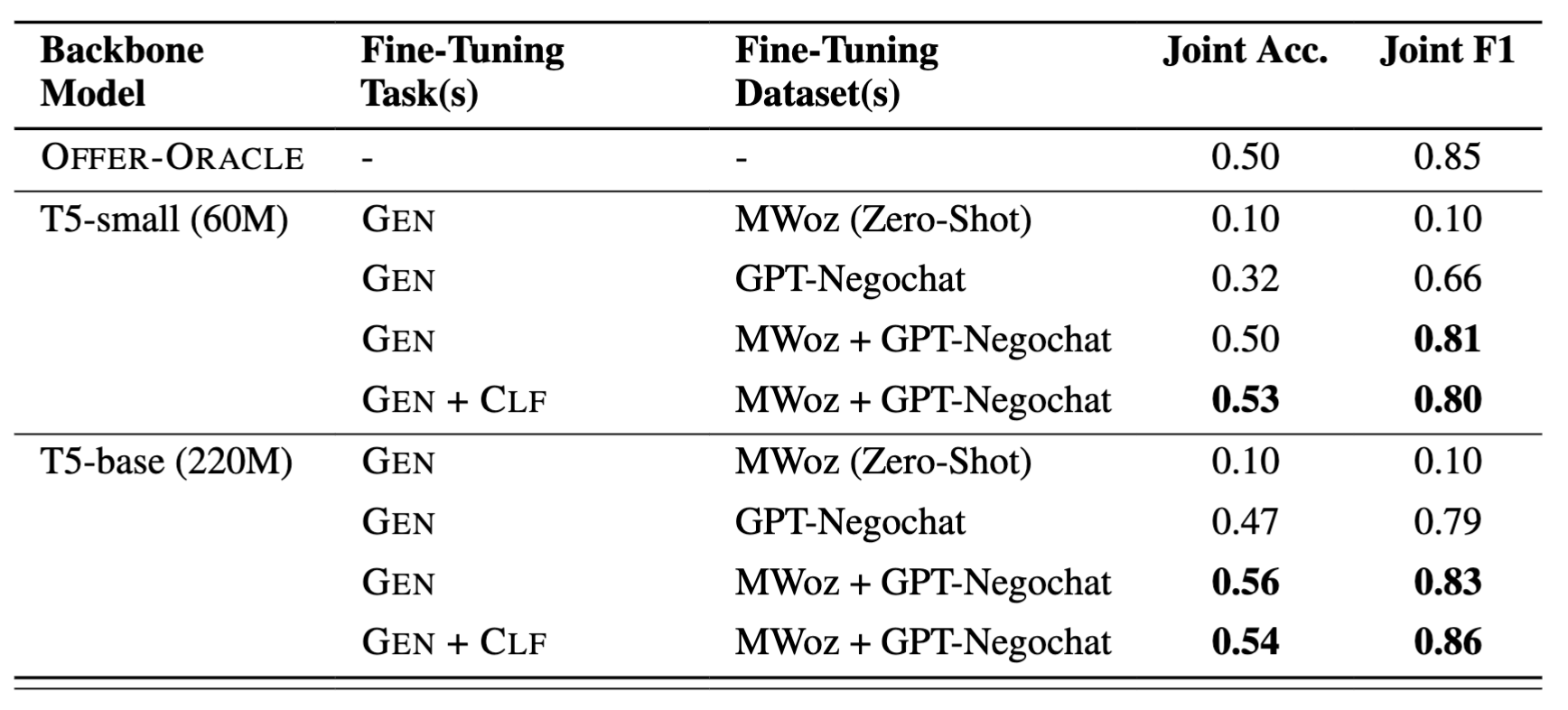}
    \caption{Results of Agreement Tracking models.\label{tab:main-results}}
\end{figure*}

\subsubsection{Multi-Task Training}
To address inaccuracies in the agreement tracking task, we borrow from multi-task training with contrastive-learning objectives in other dialogue-related tasks, e.g., summarization and state tracking \cite{chen2021improvingfaithfulness, cao2021cliffcontrastivelearning, tang-etal-2022-confit}. Our approach penalizes incorrect outputs by separating them in the model's latent space, and the model then learns to discriminate negatively-sampled \textit{A-Lev}s from the correct one.

Since we use a T5 model, a text-to-text paradigm, we implement the contrastive learning objective by incorporating an additional tokenization scheme during training. The main tokenization scheme teaches the model to predict the correct \textit{Lev} given the previous set of agreements and the dialogue context (referred to as \textsc{Gen}). We also incorporate an auxiliary tokenization scheme to meet our contrastive learning goals, providing additional training signals from which the model learns discriminative representations (referred to as \textsc{Clf}).  Here, we describe these two training tasks, along with their respective tokenization schemes. (Also, see examples in Figure \ref{fig:prompts}.)

\begin{enumerate}
    \item \textbf{\textsc{Gen}}: Our primary task focuses on conditional generation, where the model is trained to generate Levenshtein belief spans for each turn while being conditioned on: 1) the previous agreement state, and 2) the dialogue context.
    \begin{multline*}
        \mathcal{L}_{\scriptsize{\textsc{Gen}}} = -log\ P(A_t | C_t, A_{t - 1})
    \end{multline*}
    \item \textbf{\textsc{Gen + Clf}} In our multi-task setup, we supplement the primary task of agreement state prediction (\textsc{Gen}) with an auxiliary task that explicitly trains the model to discern between correct belief spans and negatively sampled incorrect outputs.
    
    \textsc{Clf} is designed as a binary classification task that takes three inputs: 1) the set of agreements as of the previous turn, 2) the dialogue context, and 3) an \textit{A-Lev}, which is randomly assigned as either the correct output or a negatively sampled variant based on the ontology. The expected output for this task is a boolean, either ``yes'' to indicate that the \textit{A-Lev} is indeed the correct output or ``no'', to indicate that the\textit{Lev} is negatively sampled.
    \begin{multline*}
        \mathcal{L}_{\scriptsize{\textsc{Gen}} + \scriptsize{\textsc{Clf}}} = -log\ P(A_t | C_t, A_{t - 1}) \\ - log\ P(Y_t | C_t, A'_{t}, A_{t - 1})
    \end{multline*}
    where $A'_{t}$ represents a randomly sampled Lev, which can either be a distractor or the correct answer. $Y_t$ is the label of our binary classification task (the binary label is tokenized in the form of ``yes'' or ``no'').
\end{enumerate}

\subsection{Evaluation Metrics}
We assess our models using Joint Slot Accuracy and Joint F1 Score, standard dialogue state tracking metrics. Joint Slot Accuracy compares predicted agreements to ground truth, requiring an exact match. Joint F1 Score, a more lenient metric, calculates precision and recall instead of using binary 0 or 1 for mispredictions.

\subsection{Training Setup}
We implement our models using HuggingFace Transformers \cite{wolf-etal-2020-transformers} and PyTorch Lightning\footnote{\href{https://github.com/Lightning-AI/lightning}{https://github.com/Lightning-AI/lightning}}. During the fine-tuning process, we use a batch size of 32 and apply early stopping based on the performance on a separate validation set. We employ the Adam optimizer \cite{kingma2014adam} with the learning rate set to $6 \times 10^{-4}$. To ensure reliable estimates, we conduct a 3-fold cross-validation scheme, averaging the results across the folds. Since sample efficiency is one of our key concerns, we run all our main experiments using different proportions of the training data: 10\%, 20\%, 30\%, 40\%, 50\%, 75\%, and 100\%. We report our hyperparameter search and best-found hyperparameter values in Appendix \ref{sec:hyperparam-appendix}.

\section{Results}
\label{sec:results}

Table \ref{tab:main-results} presents the summary of results from all the models. We elaborate on the results and make observations below.

\subsection{Rule-Based Reference Model}
Our rule-based algorithm, \textsc{Oracle-Tracker}, achieves a Joint Slot Accuracy of 0.5. This indicates that deducing agreements solely from utterance-level dialogue acts and slot-value pairs is a complex task.

\subsection{Transfer Learning from Dialogue State Tracking}
Fine-tuning T5-small and T5-base on MultiWOZ2.4 before \textsc{GPT-Negochat} improves Joint Slot Accuracy and Joint F1 score compared to fine-tuning on \textsc{GPT-Negochat} alone. Training on MultiWOZ allows the model to specialize in dialogue-related tasks, like tracking information across turns.

\paragraph{Multi-Task Training} The use of the auxiliary binary classification objective (\textsc{Gen} + \textsc{Neg}) does not lead to a significant performance improvement as compared training with the generation (\textsc{Gen}) objective alone. This result suggests that the majority of errors made by our model are \textit{false negatives}, whereas the \textsc{Neg} objective is better suited to tackle the problem of \textit{false positives}, or hallucinations. Further experiments with different contrastive learning objectives and negative sampling strategies could shed light on this hypothesis.

\begin{figure}[hptb]
    \centering
    \includegraphics[scale=0.45]{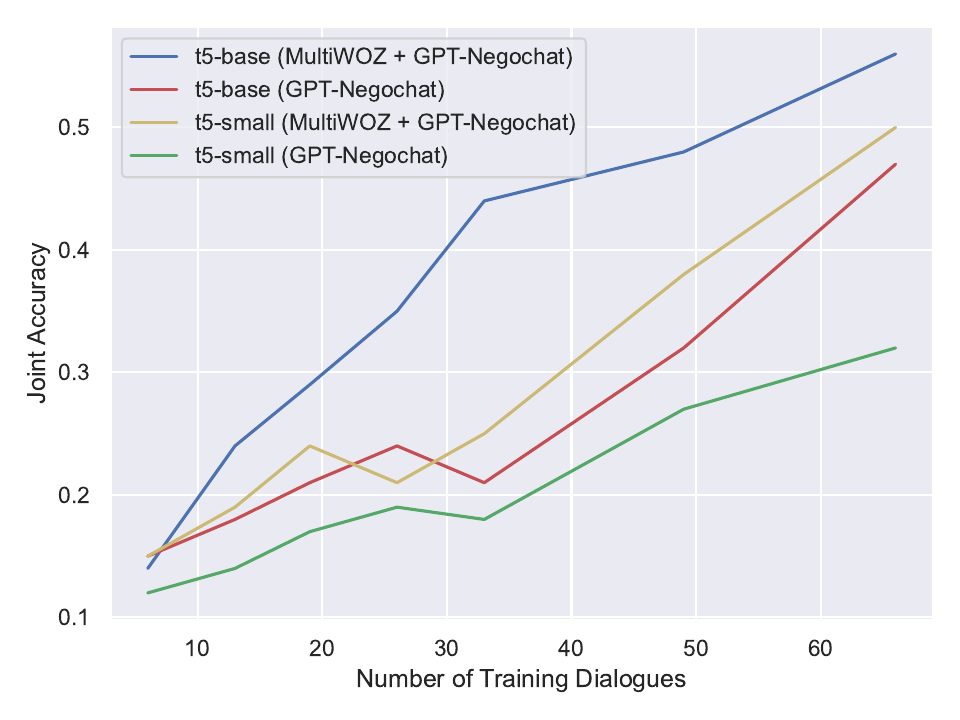}
    \caption{This plot shows the Joint Accuracy trend of four different models when trained with 10, 20, 30, 40, 50, 75, and 100 percentages of the training data. We observe positive effects on the Joint Accuracy from both model size (T5-base is larger than T5-small) as well as an additional step of fine-tuning over DST on MultiWOZ before training it on \textsc{GPT-Negochat}.}
    \label{fig:t5-vs-t5mwoz}
\end{figure}

\paragraph{Sample Efficiency} Figure \ref{fig:t5-vs-t5mwoz} illustrates the trend of Joint Slot Accuracy for models trained on different proportions of the dataset using a three-fold split. The larger T5-base model outperforms T5-small in low-resource settings. However, there is no distinct advantage of T5-base over T5-small as the training split approaches 100\%.

Our findings demonstrate that agreement tracking, although categorized as a ``dialogue state tracking'' task, cannot be adequately addressed by directly applying existing models trained on task-oriented dialogue datasets. Instead, it suggests the need for specialized techniques tailored specifically to this task.


\section{Related Work} \label{sec:related}
 
Prior to the deep-learning era, research on agreement modeling from meeting transcripts as discussed by \citet{hillard2003detectionofagreement} and \citet{galley2004identifyingagreement}, focuses on classifying dialogue acts using handcrafted features rather than tracking slot values. \citet{bui2009extractingdecisions} introduces a labeling scheme encompassing three dialogue acts: \textit{issue}, \textit{resolution}, and \textit{agreement}. \citet{frampton2009realtimedecision} investigates the impact of context window size on agreement-tracking performance. However, their model is not directly applicable to our setting, as it relies on turn-level dialogue act annotations. In contrast, our method requires only labels about the agreement state, making it more efficient and practical for the task at hand.

Recent work focuses on natural language summaries of dialogues \cite{feng2021asurveyon, gliwa2019samsumcorpusa} with factual correctness improvements potentially benefiting state tracking tasks \citet{zhao2021effectivesequence}. \citet{JiaTaxonomy} group these methods into: (1) pre-processed feature injection \cite{park-etal-2022-leveraging}, (2) design of self-supervised objectives \cite{liu-chen-2021-controllable}, and (3) using additional data \cite{liu-etal-2021-topic-aware, park-etal-2022-leveraging}.

Finally, while Transformer-based approaches have shown advancements in dialogue state tracking (DST) for task-oriented dialogues \cite{lee2021dialoguestatetracking, Peng2021SoloistBT, lin2021leveragingslotdescriptions, lin2021zeroshotdialogue}, their efficacy beyond task-oriented dialogues, such as collaborative and negotiation dialogues, remains unproven.

\section{Limitations and Future Work} \label{sec:limitations}
Although \textsc{GPT-Negochat} features a smaller ontology compared to some multi-domain DST datasets, its focus on a limited scope allows us to tackle the unique challenges posed by dynamic negotiation dialogues. Future work will seek to extend this approach to larger, multi-domain datasets. Due to computational constraints, we did not experiment with larger T5 models or instruction-tuned variants such as FLAN-T5 \cite{Chung2022ScalingIL}.
Lastly, while \textsc{GPT-Negochat} enhances the linguistic diversity of the original \textsc{Negochat} corpus, acquiring a fully organic multi-issue negotiation dialogue dataset with turn-level agreement annotations would provide a more realistic environment for evaluating different agreement tracking methods.


\section{Conclusion} \label{sec:conclusion}
This study introduces a novel task, namely agreement tracking for multi-issue negotiation dialogues, and formulates it as a variant of dialogue state tracking. We observe that fine-tuning a language model on the dialogue state tracking task for task-oriented dialogues yields improved performance over our naive baseline. The significance of state tracking extends beyond form-filling dialogues, emphasizing the ongoing need for research to improve the transferability and sample efficiency of dialogue state tracking models to other dialogue paradigms.


\section*{Broader Impact and Ethical Considerations} \label{sec:ethics}
This research introduces the task for tracking agreements in multi-issue negotiation dialogues, offering potential advancements in promote automated systems that support negotiation dialogues between human participants. A robust negotiation support system can address power imbalances and prevent exploitation or disadvantage for participants who may be in a weaker position. Even though the agreement tracker focuses on a specific task, it may still involve handling sensitive data related to the negotiation process. Researchers must prioritize proper data anonymization and take measures to protect the privacy and confidentiality of participants' information.

\bibliography{anthology,custom,refs}
\bibliographystyle{acl_natbib}

\appendix

\section{Prompt to Generate \textsc{GPT-Negochat}}
\label{sec:gpt-negochat-prompt}
The following prompt template is fed to the \textbf{text-davinci-003} variant of GPT-3 to generate rephrased utterances for \textsc{GPT-Negochat}:
\begin{quote}
    Rephrase this while still maintaining the same meaning. Feel free to add some minimal niceties and make it sound less robotic. While rejecting an offer, try to come up with a reason. Make the tone sound like a real job interview: $[\texttt{original utterance}]$
\end{quote}

\section{Sizes of Dataset Splits}
\label{sec:dataset-splits-appendix}
For all our experiments, we follow the following steps:
\begin{enumerate}
    \item We first perform a three-fold split on our entire dataset, which results in 66.67\% of the samples being assigned to the training split and the remaining to the test split.
    \item For each split, we further subdivide the training split into the actual training and the validation split using a 85\%-15\% split.
    \item Effectively, this strategy results in the training, validation, and test splits containing 56.67\%, 10\%, and 33.33\% of the dataset respectively.
\end{enumerate}

\section{Hyperparameter Tuning}
\label{sec:hyperparam-appendix}
We restrict our hyperparameter sweep to a few values over training-related hyperparameters such as learning rate, batch size, early stopping criterion, and precision. We also sweep over varying sizes of the context window for representing our dialogue context. To ensure that we do not over-optimize our hyperparameters and stumble upon ``lucky'' values, we perform an initial sweep once with a fixed configuration and apply the best-found hyperparameter values to all our other experiments.

We report the search spaces and best-found values for our hyperparameters in Table \ref{tab:hyperparams}.

\begin{table*}[ht]
\centering
\label{tab:sample-dialogue}
\begin{tabular}{l|l|l}
\multicolumn{1}{c|}{\textbf{Hyperparameter}} & \multicolumn{1}{c}{\textbf{Sweep Range}} & \textbf{Best Value} \\ 
\toprule
Learning Rate & 1 $\times 10^{-4}, 1 \times 10^{-5}, 1 \times 10^{-3}, 6 \times 10^{-4}$ & $6 \times 10^{-4}$ \\ 
\hline
Batch Size & 32, 64, 128 & 32 \\ 
\hline
Early Stopping (Min Delta) & 0, 0.001, 0.005, 0.1, 0.5 & 0 \\ 
\hline
Early Stopping (Patience) & 0, 1, 2, 3, 4 & 4 \\ 
\hline
Gradient Clip Norm & 1.0 & 1.0 \\ 
\hline
Precision & 16, 32 & 32 \\ 
\hline
Context Window Size & 1, 2, 3, 4 & 3 \\
\hline
\end{tabular}
\caption{Hyperparameter values for all experiments.\label{tab:hyperparams}}
\end{table*}
\section{A (Longer) Side-by-Side Comparison of Utterances}
Table \ref{tab:nego-vs-gpt-longer} shows an additional list of side-by-side examples of original and rephrased utterances in the \textsc{Negochat} and \textsc{GPT-Negochat} datasets respectively.

\label{sec:gpt-neg-vs-neg-examples}
\begin{table*}[ht]
\centering
\resizebox{1\textwidth}{!}{
\small
\begin{tabular}{@{}p{2cm}p{6cm}p{6cm}@{}}
\toprule
\textbf{Speaker} & \textbf{Original Utterance} & \textbf{Rephrased Utterance} \\ \midrule
Candidate & I would like a workday of 8 hours & I'm looking for something that would require 8 hours a day. \\
Employer & cand you do 10? & Can you do 10 hours a day? \\
Candidate & I disagree about a workday of 10 hours. I would like a workday of 8 hours & I'm not comfortable with that. I'd prefer 8 hours a day. \\
Employer & what else pension, car? & What else do you need, like a pension or a car? \\ \hline
Candidate & I accept programmer position. Now let's talk about how we can agree on other issues. I want a salary of 120,000? & I'd like to accept the programmer position. How about we discuss the other details, such as salary? I was hoping for something around 120,000. \\ 
Employer & I would like to offer you 90,000 for salary. & That's a bit higher than what we had in mind. We were thinking of offering 90,000. \\ 
Candidate & I would like a pension of 20\% & Could I get a pension of 20\%? \\ \hline
Candidate & I accept 60,000. I would like to continue and discuss another issue. I want a position of project manager & Yes, that's fine. I'd also like to discuss another topic - I'm hoping for a project manager position. \\
Employer & I accept project manager. & Absolutely, we can make that happen. \\
Candidate & It's good to reach an agreement on project manager & It's great that we can come to an agreement on the project manager. \\
Candidate & I want a company car & I want a company car \\
Employer & I do not accept leased car, can you manage without one? & I do not accept leased car, can you manage without one? \\
Candidate & I reject no company car. I would like a company car & I definitely want a company car. \\ \hline
\bottomrule
\end{tabular}
}
\caption{More excerpts from the \textsc{Negochat} corpus alongside its rephrased counterparts from \textsc{GPT-Negochat}.\label{tab:nego-vs-gpt-longer}}
\end{table*}

\section{Instructions for Human Annotators} \label{sec:human-annotation-instructions}
\paragraph{Meaning.} The key thing we are interested in is to see whether the paraphrased utterance alters the FUNDAMENTAL meaning of the original utterance in the negotiation dialogue.
Therefore, some examples of MEANING ALTERING utterances would be:\\\\
A: No, I'd like a product manager position\\
B: Yes, I'd like a product manager position\\
A: I can only work 8 hours\\
B: I can only work 9 hours\\\\
Similarly, some examples of utterances that DO NOT ALTER MEANING would be:\\
Context: We can certainly make that happen with a 10\% pension fund.\\
A: No, I don't think that's the right fit for me.\\
B: I disagree\\
Context: 9 hours sounds good to me. Now, I'm asking for a 20\% pension.\\
A: I can agree to that pension.\\
B: Alright, I can work with that.\\
\paragraph{Naturalness.} In addition to judging the same underlying meaning/effect of the utterance, you will also be asked to pick which of the two utterances (if any) sound more realistic in a real-world job negotiation. For example, it's unlikely that a professional interview candidate would say something super casual like "Nah man, you can do better than 80K".

\end{document}